\documentclass[conference]{IEEEtran}
\IEEEoverridecommandlockouts
\usepackage[noadjust]{cite}
\usepackage{amsmath,amssymb,amsfonts,nccmath,bm}
\usepackage{textcomp}
\usepackage{xcolor}
\usepackage[ruled,linesnumbered,vlined]{algorithm2e}
\usepackage{graphicx}
\graphicspath{ {images/} }
\usepackage{caption}
\usepackage{subcaption}
\usepackage{url}
\usepackage{microtype}
\usepackage{bm}
\usepackage{booktabs}
\usepackage[hidelinks]{hyperref}
\usepackage[skip=1pt]{caption}
\usepackage{cleveref}

\def\BibTeX{{\rm B\kern-.05em{\sc i\kern-.025em b}\kern-.08em
    T\kern-.1667em\lower.7ex\hbox{E}\kern-.125emX}}

\definecolor{darkgreen}{rgb}{0.0, 0.2, 0.13}

\begin{document}

\title{Temporal Cascade and Structural Modelling of EHRs for Granular Readmission Prediction}

\author{
\IEEEauthorblockN{Bhagya Hettige}
\IEEEauthorblockA{\textit{Monash University}\\
\textit{bhagya.hettige@monash.edu}}
\and
\IEEEauthorblockN{Weiqing Wang}
\IEEEauthorblockA{\textit{Monash University}\\
\textit{teresa.wang@monash.edu}}
\and
\IEEEauthorblockN{Yuan-Fang Li}
\IEEEauthorblockA{\textit{Monash University}\\
\textit{yuanfang.li@monash.edu}}
\and
\IEEEauthorblockN{Suong Le}
\IEEEauthorblockA{\textit{Monash Health}}
\textit{suong.le@monashhealth.org}
\and
\IEEEauthorblockN{Wray Buntine}
\IEEEauthorblockA{\textit{Monash University}\\
\textit{wray.buntine@monash.edu}}
}

\maketitle

\begin{abstract}
Predicting (1) \emph{when} the next hospital admission occurs and (2) \emph{what} will happen in the next admission about a patient by mining electronic health record (EHR) data can provide granular readmission predictions to assist clinical decision making. 
Recurrent neural network (RNN) and point process models are usually employed in modelling temporal sequential data. 
Simple RNN models assume that sequences of hospital visits follow strict causal dependencies between consecutive visits.
However, in the real-world, a patient may have multiple co-existing chronic medical conditions, i.e., multimorbidity, which results in a \emph{cascade} of visits where a non-immediate historical visit can be most influential to the next visit. 
Although a point process (e.g., Hawkes process) is able to model a cascade temporal relationship, it strongly relies on a prior generative process assumption.
We propose a novel model, \texttt{MEDCAS}, to address these challenges. 
\texttt{MEDCAS} combines the strengths of RNN-based models and point processes by integrating point processes in modelling visit types and time gaps into an attention-based sequence-to-sequence learning model, which is able to capture the temporal cascade relationships. 
To supplement the patients with short visit sequences, a structural modelling technique with graph-based methods is used to construct the markers of the point process in \texttt{MEDCAS}. 
Extensive experiments on three real-world EHR datasets have been performed and the results demonstrate that \texttt{MEDCAS} outperforms state-of-the-art models in both tasks. 

\end{abstract}




\section{Introduction}

The rapid increase of Electronic health record (EHR) data collected at hospitals presents an opportunity for applying machine learning techniques for healthcare analysis~\cite{DBLP:journals/titb/deep_ehr,goldstein2017opportunities,DBLP:journals/csur/ehr_survey}.
EHRs form longitudinal sequences of patient visits.
Each visit records a set of medical codes (e.g., diagnosis, procedure and medication codes) observed during the visit.
Sequential disease prediction, which predicts \emph{what} medical codes will be observed in the next visit, 
is a core task in personalised healthcare. 
In addition, the knowledge of \emph{when} a patient's next visit might occur allows hospitals to better allocate resources (e.g., beds, staff) and proactively reduce preventable readmissions~\cite{journal/jama/readmission}.
Thus, the ability to make accurate predictions for both \emph{when} and \emph{what} aspects of future visits is important for personalised healthcare, which can produce \emph{granular readmission predictions}.
To the best of our knowledge, only a few studies simultaneously model both tasks~\cite{DBLP:conf/mlhc/doctorai,DBLP:conf/ijcai/pacrnn}. This work is motivated by the observation that
EHRs contain rich clinical information that can facilitate both tasks.

\begin{figure}[!t]
    \begin{subfigure}{\linewidth}
        \centerline{\includegraphics[width=\columnwidth]{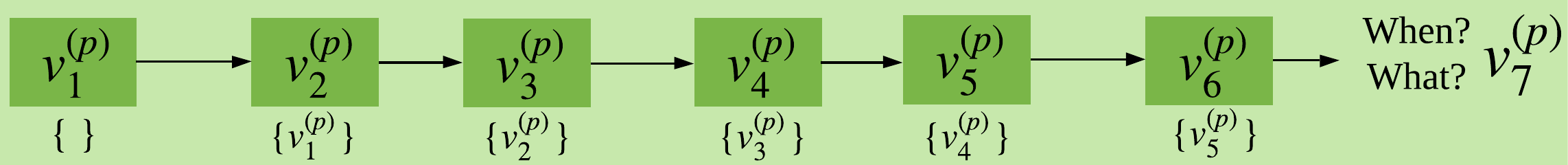}}
        \caption{Visit sequence with strict consecutive dependencies.}
        \label{fig:seq_visits}
    \end{subfigure}

    \begin{subfigure}{\linewidth}
        \centerline{\includegraphics[width=\columnwidth]{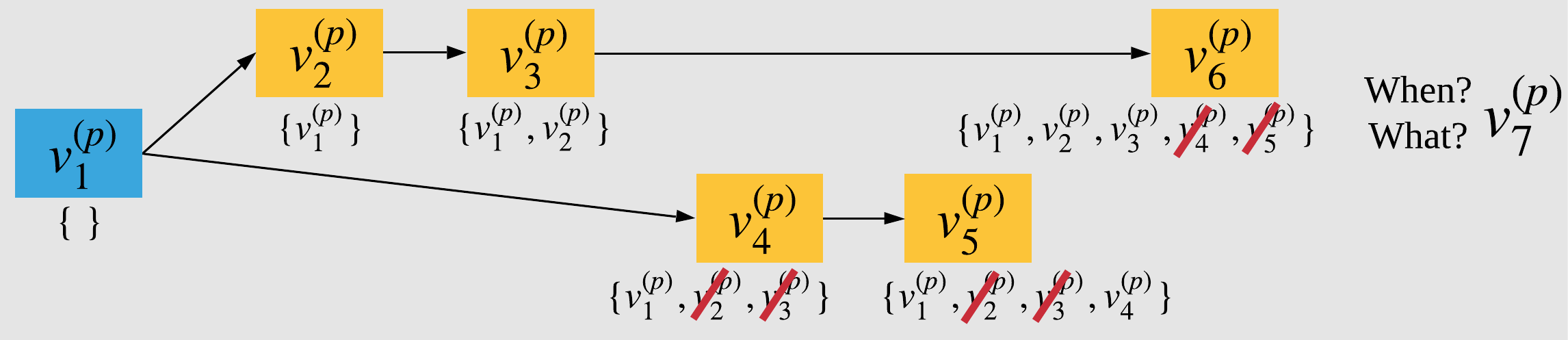}}
        \caption{Temporal cascade visit sequence.}
        \label{fig:cas_visits}
    \end{subfigure}
    
    \caption{Deconstructing the strict temporal dependencies between consecutive visits. Ancestor visits are shown within brackets.}
    \label{fig:cascade_visits}
\end{figure}

Most existing EHR modelling studies~\cite{DBLP:conf/mlhc/doctorai,arxiv/medgraph} assume that a patient's visit sequence follows strict causal dependencies between consecutive visits, so that each visit is influenced by its immediately preceding visit, as shown in Figure~\ref{fig:seq_visits}. 
These techniques employ a chain structure to capture the memory effect of past visits.
Prior research indicates that \textbf{multimorbidity}~\cite{almirall2013coexistence}, which is defined as the co-occurrence of two or more chronic conditions, has moved onto the priority agenda for many health policymakers and healthcare providers \cite{journal/Navickas/Multimorbidity,journal/Afshar/Multimorbidity}. According to a global survey 
\cite{journal/Afshar/Multimorbidity}, the mean world multimorbidity prevalence for low and middle-income countries was 7.8\%. Also, in Europe, patients with multimorbidity are 
some of the most costly and difficult-to-treat patients \cite{journal/Navickas/Multimorbidity}. 
This prevalent and costly phenomenon induces complex causal relationships between visits where a non-immediate historical visit can be most influential 
to the next visit, as shown in Figure~\ref{fig:cas_visits}.
We term this effect
``\textbf{temporal cascade relationships}''.

Attention based RNN models, such as Dipole~\cite{DBLP:conf/kdd/dipole} and RETAIN~\cite{DBLP:conf/nips/retain}, are capable of modelling these temporal cascade relationships. However, they both regard visit sequences as evenly spaced time series and overlook the influence of time gaps between visits which is usually not the case in real-world medical data. Naturally, they cannot predict \emph{when}. 
In contrast, temporal point processes naturally describe events that occur in continuous time. 
Typical point process models, such as Hawkes~\cite{hawkes_process}, strongly rely on prior knowledge of the data distribution and the excitation process with strict parametric assumptions, which is usually not observed in the real-world. 
Neural point process models, such as RMTPP~\cite{DBLP:conf/kdd/rmtpp}, capture these unknown temporal dynamics using RNN. 
Yet, they mimic the 
chained structure of event sequences to model historical memory effect, and are not capable of modelling the cascade relationships. 

We propose a novel model, \texttt{MEDCAS}, to address these problems. 
\texttt{MEDCAS} combines the strengths of point process models and attention-based RNN models, and is able to capture the cascade relationships with time gap influences without a strong prior data distribution assumption. 
The basic idea of \texttt{MEDCAS} is to integrate point processes in learning visit type and time gap information into an attention-based RNN framework. 
Different from existing attention-based RNN models (e.g., Dipole~\cite{DBLP:conf/kdd/dipole} and RETAIN~\cite{DBLP:conf/nips/retain}) where attention is optimized for predicting \textit{what} only, \texttt{MEDCAS} adopts a \emph{time-aware} attention-based RNN encoder-decoder structure where 
the attention is learned not only in predicting \emph{what}, but also in modelling the intensity to predict \emph{when}.
Thus, the cascade relationships for visit sequences captured by \texttt{MEDCAS} are also \emph{time-aware} compared to existing EHR models where time is usually ignored. 
Moreover, a novel intensity function is introduced, so that no prior knowledge of the data and the excitation process is required for modelling cascade relationships, compared with existing point process models. 

\begin{figure}[!t]
    \begin{subfigure}{0.49\columnwidth}
        \centering
        \includegraphics[width=0.8\columnwidth]{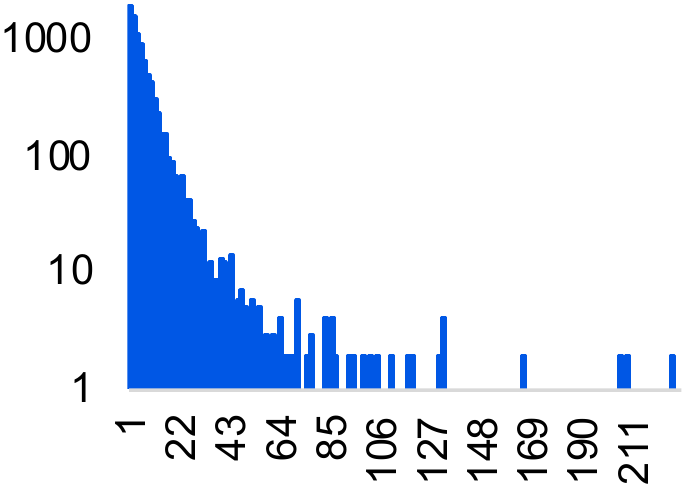}
        \caption{HF}
    \end{subfigure}
    \begin{subfigure}{0.49\columnwidth}
        \centering
        \includegraphics[width=0.7\columnwidth]{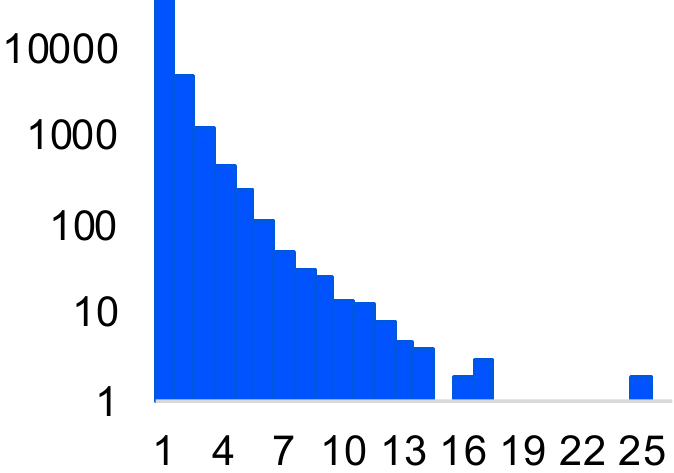}
        \caption{MIMIC}
    \end{subfigure}
    
    \caption{$log_{10}(\text{\# patients})$
    against visit sequence lengths.
    }
    \label{fig:seq_lens_distribution}
\end{figure}

\begin{figure}[!t]
    \centering
    \includegraphics[width=0.6\columnwidth]{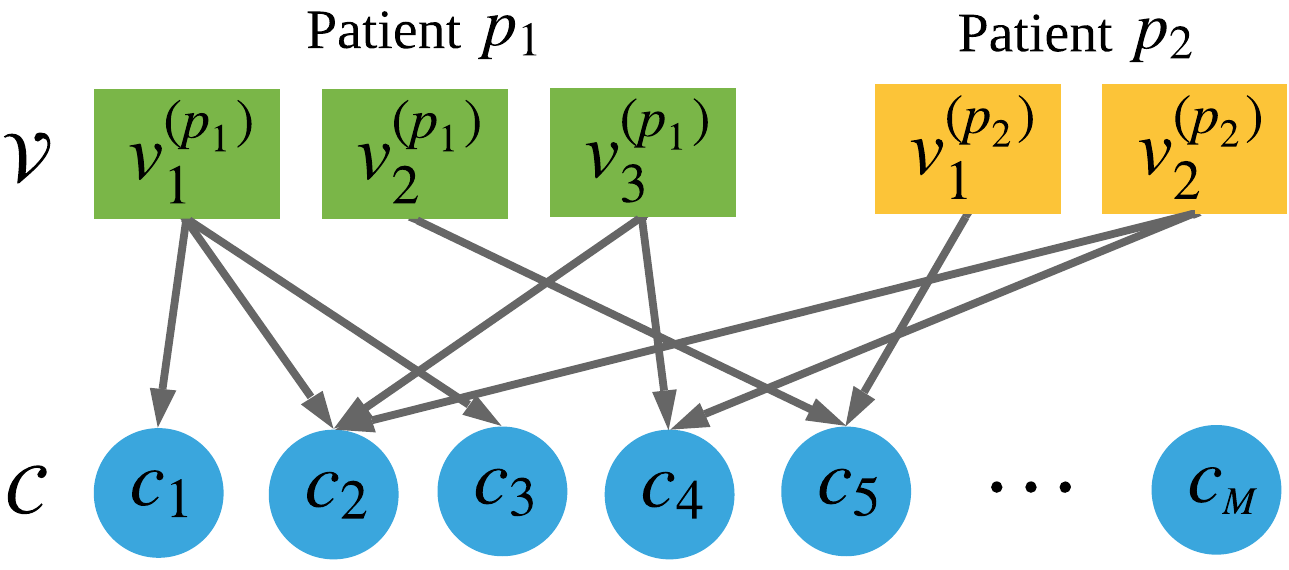}
    \caption{Code-sharing behavior of visits.}
    \label{fig:struct_vc}
\end{figure}

Many patients in EHR data have short medical histories, as shown in Figure~\ref{fig:seq_lens_distribution} for two real-world datasets used in our experiments.
Both distributions are highly imbalanced, with a median of 5 visits in HF and 1 visit in MIMIC.
We term this problem the ``\textbf{cold start patient problem}'' in the healthcare domain. 
Inspired by the structural modelling technique proposed in MedGraph~\cite{arxiv/medgraph}, \texttt{MEDCAS} encodes the visit-code relationships (Figure~\ref{fig:struct_vc}) to learn the local (within a patient) and global (across other patients) code-sharing behaviours of visits. 
With the globally captured visit similarities in terms of medical conditions across other patients, we can effectively alleviate the cold start patient problem in the absence of much temporal data about a patient.

The main contributions of this paper are threefold: 
\begin{enumerate}
    \item A granular readmission prediction model, \texttt{MEDCAS}, which (1) simultaneously predicts \emph{when} the next visit occurs and \emph{what} happens in the next visit (2) effectively models multimorbidity phenomenon without requiring prior knowledge about the EHR data distribution. 
    

    \item A novel architecture capable of modelling \textit{time-gap aware cascade relationships} in patients' visit sequences by integrating point process into an attention-based RNN encoder-decoder structure where the attention is learned for predicting not only \textit{what} but also \textit{when}. To generate visit markers for the point process effectively especially for the cold-start patients, a graph-based structure modelling method is adopted and a novel sampling method has been proposed to avoid cluttered joint training of structural and temporal parts accordingly.
    
    \item An extensive experimental study conducted on three real-world EHR datasets that demonstrates the superiority of \texttt{MEDCAS} over state-of-the-art EHR models. 
\end{enumerate}

\section{Related Work}

Over the years, many EHR modelling methods~\cite{DBLP:journals/titb/deep_ehr,goldstein2017opportunities,DBLP:conf/icdm/GaoWWYGWT019,gao2019camp,kddd_2020,ehr_icdm19,kdd_2020_risk} have been proposed to leverage the patterns of patients, visits and codes to answer questions on predicting various medical outcomes of patients, such as next disease prediction~\cite{DBLP:conf/kdd/dipole,DBLP:conf/ijcai/pacrnn,DBLP:conf/icdm/GaoWWYGWT019,gao2019camp,kddd_2020}, medical risk prediction~\cite{DBLP:conf/kdd/med2vec,ehr_icdm19,kdd_2020_risk} and mortality prediction~\cite{arxiv/medgraph,conf/aaai/GCT}. 

Med2Vec~\cite{DBLP:conf/kdd/med2vec}, a Skip-gram-based method~\cite{DBLP:conf/nips/word2vec}, learns to predict codes of the future visit without temporal dependencies. 
Dipole~\cite{DBLP:conf/kdd/dipole} is an attention-based bi-directional RNN model that captures long-term historical visit dependencies for the prediction of the next visit's codes.
DoctorAI~\cite{DBLP:conf/mlhc/doctorai}, also based on RNN, captures time gaps along with a disease prediction task, but still uses the chained structure of the visit sequences.
RETAIN~\cite{DBLP:conf/nips/retain} is an interpretable predictive model with reverse-time attention.
T-LSTM~\cite{DBLP:conf/kdd/tlstm} implements a time-aware LSTM cell to capture the time gaps between visits.
MiME~\cite{DBLP:conf/nips/mime} models the hierarchical encounter structure in the EHRs.
But, the hierarchical granularity of medical codes is not always available in the real-world, thus it has limited usage. 
MedGraph~\cite{arxiv/medgraph}, exploits both structural and temporal aspects in EHRs which uses a point process model~\cite{DBLP:conf/kdd/rmtpp} to capture the time gap irregularities, but fails to model the cascade relationships.
GCT~\cite{conf/aaai/GCT} models a graph to learn the implicit code relations using self-attention transformers, but ignores the temporal aspect of visits.
ConCare~\cite{DBLP:conf/aaai/concare} is another recent work which uses a transformer network to learn the temporal ordering of the visits.
Some prominent baselines are compared in Table~\ref{tab:baselines}.

\begin{table*}[tb]
    \begin{center}
    \caption{Comparison with existing work (abbreviations: \textbf{ATTN} - attention on the past and \textbf{PP} - point process modelling).}
    \label{tab:baselines}
    \small
    \begin{tabular}{lcccccc}
        \toprule
        \textbf{EHR Modelling Method} & \textbf{Technique} & \textbf{Temporal} & \textbf{Cascades} & \textbf{Irregular time} & \textbf{Structural} & \textbf{Time $+$ Disease}\\
        \midrule
        Med2Vec~\cite{DBLP:conf/kdd/med2vec} & Skip-gram & $\times$ & $\times$ & $\times$ & \checkmark & $\times$ \\
        DoctorAI~\cite{DBLP:conf/mlhc/doctorai} & RNN & \checkmark & $\times$ & $\times$ & $\times$ & \checkmark \\
        RETAIN~\cite{DBLP:conf/nips/retain} & RNN$+$ATTN & \checkmark & \checkmark & $\times$ & $\times$ & $\times$ \\
        Dipole~\cite{DBLP:conf/kdd/dipole} & RNN$+$ATTN & \checkmark & \checkmark & $\times$ & $\times$ & $\times$ \\
        T-LSTM~\cite{DBLP:conf/kdd/tlstm} & RNN & \checkmark & $\times$ & \checkmark & $\times$ & $\times$ \\
        MedGraph~\cite{arxiv/medgraph} & RNN$+$PP & \checkmark & $\times$ & \checkmark & \checkmark & $\times$ \\
        \texttt{MEDCAS} (ours) & RNN$+$PP$+$ATTN & \checkmark & \checkmark & \checkmark & \checkmark & \checkmark \\
        \bottomrule
    \end{tabular}
    \end{center}
    \vspace{-4mm}
\end{table*}

Several recent studies~\cite{arxiv/medgraph,DBLP:conf/ijcai/pacrnn} have explored visit sequence modelling with marked point processes.
Deep learning approaches~\cite{DBLP:conf/kdd/rmtpp,DBLP:conf/nips/neur_tpp} have been developed to model the marked point processes which maximises the likelihood of observing the marked event sequences.
CYAN-RNN~\cite{DBLP:conf/ijcai/cascade_dyn} proposed a sequence-to-sequence approach to model the cascade dynamics in information diffusion networks, such as social-media post sharing.

\section{Methodology}

\subsection{Problem Definition}

The input data is a set of visit sequences $\{\mathcal{S}^{(1)}, \mathcal{S}^{(2)}, \ldots, \mathcal{S}^{(P)} \}$ for $P$ patients. 
For a patient $p$, his/her visit history denoted as $\mathcal{S}^{(p)} = \{v^{(p)}_i = (t^{(p)}_i, \mathbf{x}^{(p)}_i)\}_{i=1}^{N^{(p)}}$
consists of a temporally ordered sequence of tuples of $N^{(p)}$ finite timesteps with $i \in \mathbb{Z}^{+}$,
where $v^{(p)}_i$ is the $i$-th visit of patient $p$, in which
$t^{(p)}_i \in [0, +\infty)$ is the visit's timestamp, and $\mathbf{x}^{(p)}_i$ denotes the set of medical codes (i.e., diagnosis and procedure codes) associated with the visit.
$\mathbf{x}^{(p)}_i \in \{0, 1\}^{|\mathcal{C}|}$ is a multi-hot binary vector which marks the occurrence of medical codes from taxonomy $\mathcal{C} = \{c_1, c_2, \dots, c_{|\mathcal{C}|}\}$ in the visit.
Given a patient $p$'s visit history up to timestep $t^{(p)}_i$, our task is to answer the following questions about $p$'s next visit, $v^{(p)}_{i+1}$:
\begin{enumerate}
    \item \emph{when} will the next visit happen?
    \item \emph{what} (i.e., medical codes) will happen in the next visit?
\end{enumerate}

\subsection{\texttt{MEDCAS} Architecture}

\begin{figure}[t]
    \centering
    \includegraphics[width=1\linewidth]{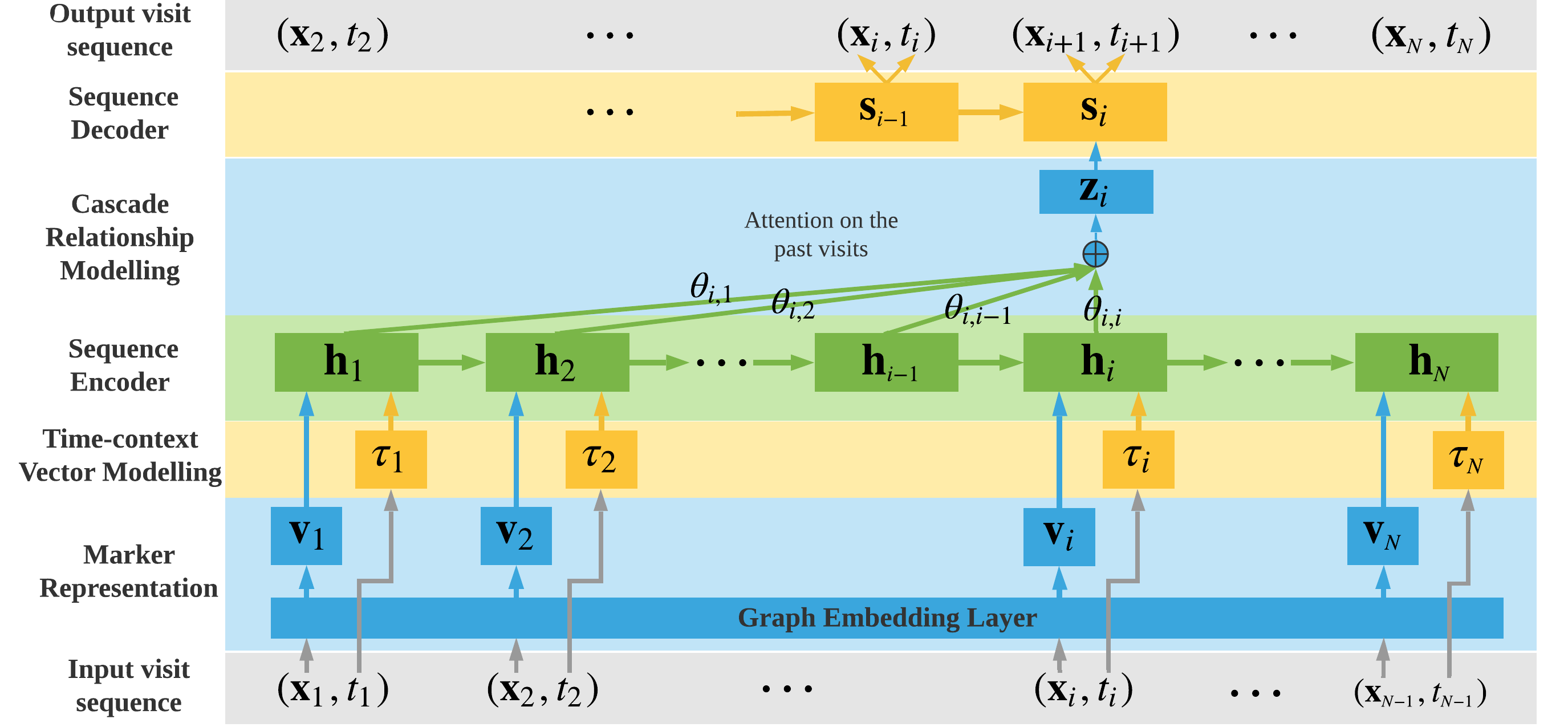}
    \caption{The \texttt{MEDCAS} architecture.}
    \label{fig:archi}
    \vspace{-6mm}
\end{figure}

Without loss of generality, we present \texttt{MEDCAS} for a single patient.
The overall architecture of \texttt{MEDCAS} is shown in Figure~\ref{fig:archi}.
Given a patient's visit sequence from timestep $1$ to $i$, our task is to predict the \emph{time} and \emph{diseases} of the next visit, i.e., $(t_{i+1}, \mathbf{x}_{i+1})$. 
Using this input visit sequence, we model a point process to capture the time gap influences of the past visits.
First, we transform each $\mathbf{x}_{i}$ into a latent visit vector, $\mathbf{v}_i\in\mathbb{R}^{D_m}$, by exploiting the graph structural visit-code relationships in EHR, shown in Figure~\ref{fig:struct_vc}. 
This $\mathbf{v}_i$ will be used as the marker in the point process.
Then, we construct a time-context vector, $\bm{\tau}_i$, with a linear transformation to effectively model the time gap irregularities between consecutive visits.
Using the marker $\mathbf{v}_i$ and the time-context $\bm{\tau}_i$ vectors, we model the temporal \emph{cascade relationships} through a time-aware attention-based encoder-decoder alignment model.
At the $i$-th timestep, the alignment model predicts the time and the diseases of the next visit $(t_{i+1}, \mathbf{x}_{i+1})$.
To align the input and the output visit sequences, we employ an aggregated attention-based ancestor-context vector $\mathbf{z}_i$, which computes the attention of each historical visit on the next visit.
This ancestor-context vector models the implicit cascade relationships.
Accordingly, we define the intensity function which models the temporal cascade relationships of the visit sequences using both causal and implicit temporal relationships.
Since we jointly optimise ancestor visit alignment and the point process intensity, 
the modelled temporal cascade relationships learned via attention are \emph{time-informed}.

\subsubsection{Marker Representation}
Existing point process modelling methods~\cite{DBLP:conf/kdd/rmtpp,DBLP:conf/ijcai/cascade_dyn} capture sequences of events with a one-hot marker, unsuitable to represent visits which typically have multiple medical codes, i.e., high-dimensional, sparse multi-hot vectors $\mathbf{x}_i$.

\textbf{Graph embedding layer:}
A na\"{i}ve approach to deal with simultaneous event types is to encode the multi-hot codes into a latent representation before feeding into the temporal model.
However, this simple encoding is unable to capture the complex \emph{local} (codes connected to a visit) and \emph{global} (visits sharing similar codes) code-sharing behaviours of visits. 
Inspired by the bipartite graph technique proposed in MedGraph~\cite{arxiv/medgraph} to capture visit-code relationships, we construct an unweighted bipartite graph $G = (\mathcal{V} \cup \mathcal{C}, E)$, where $\mathcal{V} = \{v_1, v_2, \dots v_{N}\}$ is the visit node partition, $\mathcal{C}$ is the code node partition, and $E$ is the set of edges between them
(Figure~\ref{fig:struct_vc}).
$E$ can be constructed from the multi-hot code vector $\mathbf{x}_i$ for each visit node $v_i$. 

Attributed nodes enable inductive inference of future/unseen visits~\cite{g2g}.
For visit $v_i$'s attribute vector, we use $\mathbf{x}_i$ which denotes the codes in $v_i$, since it can uniquely distinguish each visit.
A one-hot vector is used for representing code nodes as we do not use supplementary code data.
i.e., $\mathbf{m}_j \in \mathbf{I}_{|\mathcal{C}|}$ where $\mathbf{I}_{|\mathcal{C}|}$ is the identity matrix, is used as the attribute vector of each code $c_j$. 
We use an attributed graph embedding method to learn the visit-code relationships to obtain visit and code embeddings.
In this work, we use the second-order proximity~\cite{DBLP:conf/www/line} to capture visit neighbourhood similarity.

Let $(v_i, c_j) \in E$ be an edge between visit $v_i$ and code $c_j$ with attributes $\mathbf{x}_i \in \{0, 1\}^{|\mathcal{C}|}$ and $\mathbf{m}_j \in \{0, 1\}^{|\mathcal{C}|}$, respectively.
We project the two node types to a common semantic latent space using two transformation functions parameterised by $\mathbf{W}_v, \mathbf{W}_c \in \mathbb{R}^{D_m \times {|\mathcal{C}|}}, \mathbf{b}_v, \mathbf{b}_c \in \mathbb{R}^{D_m}$, 
where $D_m$ is the dimension of the latent vector.
\begin{equation}\small
    \mathbf{v}_i = \mathbf{W}_v^T \mathbf{x}_i + \mathbf{b}_v \; \text{ and } \; \mathbf{c}_j = \mathbf{W}_c^T \mathbf{m}_j + \mathbf{b}_c
    \label{eq:u_transform}
\end{equation}

We define the probability that the patient is diagnosed with disease $c_j$ during hospital visit $v_i$ as the conditional probability of code $c_j$ generated by visit $v_i$ as: $p(c_j | v_i) = \frac{\exp (\mathbf{c}_j^\top \cdot \mathbf{v}_i)}{\sum_{c_{j\prime} \in \mathcal{C}} \exp (\mathbf{c}_{j\prime}^\top \cdot \mathbf{v}_i)}$.
For each visit $v_i \in \mathcal{V}$, $p(\cdot | v_i)$ defines the conditional distribution over all the codes in $\mathcal{C}$, so that the visit's neighbourhood is modeled by this formulation.
We define the empirical probability of observing code $c_j$ in visit $v_i$ as $\hat{p}(c_j | v_i) = \frac{1}{degree (v_i)} = \frac{1}{sum(\mathbf{x}_i)}$.
Therefore, to preserve the second-order proximity for visit node $v_i$, the conditional probability $p(\cdot | v_i)$ and the empirical probability $\hat{p}(\cdot | v_i)$ need to be closer. 
By applying Kullback Leibler (KL) divergence and omitting the constants~\cite{DBLP:conf/www/line}, we can derive the structural loss for the visit-code graph as:
\begin{equation}\small
    \label{eq:L_struct}
    \mathcal{L}_{s} = -\sum_{(v_i,c_j) \in E} \!\!\!\! \log \ p(c_j|v_i)
\end{equation}

We use the visit embedding $\mathbf{v}_i$, which captures the code-sharing behaviours, as the latent marker of the $i$-th visit.

\subsubsection{Time-context Vector Modelling}

To model the point process, we need to define the time-context of events.
Accordingly, for each visit we consider its time gap from the immediately previous visit. 
Since time gaps are irregular and spans from hours to years in EHR systems, 
we transform this time gap to a log scaled value, so for timestep $i$, ${\Delta}_i = \log (t_i - t_{i-1})$. 
Existing approaches simply concatenate this scalar time value to the marker vector. 
However, when the event marker vector is high-dimensional, it is likely that the model will ignore the effects of a scalar time-context value, which has been experimentally validated in the ablation analysis. 
Thus, we convert this scalar into a high-dimensional vector to capture the time gap irregularities more effectively. 
We use a linear transformation, due to its simplicity and experimentally validated effectiveness, to construct time gap vectors $\bm{\tau}_i = \mathbf{W}_{\Delta} \cdot {\Delta}_i \in \mathbb{R}^{D_t}$, where $D_t$ is the time-context vector dimension.
The linear transformation $\mathbf{W}_{\Delta} \in \mathbb{R}^{D_t \times 1}$ is initialised with a 0-mean Gaussian.

\subsubsection{Sequence Encoder}
The encoder reads the input visit sequence up to timestep $i$.
\texttt{MEDCAS} constructs the hidden state vector of the timestep $i$ using an RNN with $D$ units as:
\begin{equation}\small
    \mathbf{h}_i = RNN(\mathbf{h}_{i-1}, [\mathbf{v}_i, \bm{\tau}_i])
    \label{eq:seq_enc_h}
\end{equation}
Intuitively, the sequence encoder captures the history of the patient in a chained structure. 
For each timestep $i$, we concatenate the marker vector $\bm{v}_i$ and the time-context vector $\bm{\tau}_i$ as input to the RNN along with the previous hidden state. 
As a result, $\mathbf{h}_i$ carries \emph{time-aware memory} on the patient's medical disease progression up to timestep $i$.

\subsubsection{Cascade Relationship Modelling}

\texttt{MEDCAS} identifies the candidate ancestor visits at timestep $i$ using an attention-based scheme.
The ancestor visits are aggregated into an ancestor-context vector $\mathbf{z}_i$ weighted by their attention.
Different from NMT models, we restrict that only the visits up to $i$ are considered as ancestors of the next visit at $i+1$.
\begin{equation}\small
    \mathbf{z}_i = \sum_{j=1}^i \theta_{i,j} \cdot \mathbf{h}_j
    \label{eq:context_parent}
\end{equation}
where $\theta_{i,j} \geq 0$ with $\sum_{j=1}^i \theta_{i,j} = 1$ is the likelihood of the visit $v_j$ being an influential visit for the next visit.
Since the output of timestep $i$ is $v_{i+1}$ in the decoder, this gives the likelihood of $v_j$ being an ancestor visit of the visit $v_{i+1}$ in the cascade relationship setting.
The attention score is defined with the decoder hidden state which is used to align the input visit sequence to the output visit sequence.

\subsubsection{Sequence Decoder}

In the RNN decoder with $D$ units, we feed the previous decoder hidden state $\mathbf{s}_{i-1}$, encoder input of current timestep $\mathbf{x}_i$ (i.e., the teacher forcing learning scheme) and ancestor-context vector $\mathbf{z}_i$ to construct the \emph{latent cascading state} at the current timestep $\mathbf{s}_i$ as:
\begin{equation}
    \mathbf{s}_i = RNN(\mathbf{s}_{i-1}, \mathbf{x}_i, \mathbf{z}_i)
\end{equation}

We define an alignment score for each historical visit at timestep $j$ and the output visit at timestep $i$ (since we let the output sequence to be one timestep ahead of the input sequence, output at timestep $(i+1)$ is the input at timestep $i$) as: $\mathbf{e}_{i,j} = a(\mathbf{s}_{i-1}, \mathbf{h}_j)$,
where $a$ is a feedforward network 
which is jointly trained with the complete model.
The attention weight to model the cascade relationships is defined as:
\begin{equation}\small
    \theta_{i,j} = \frac{\exp (\mathbf{e}_{i,j})}{\sum_{k=1}^{i} \exp (\mathbf{e}_{i,k})}
    \label{eq:theta_ij}
\end{equation}

\subsubsection{Output Visit Sequence}
The two tasks of \texttt{MEDCAS} are to predict, (1) \emph{when} will the $(i+1)$-th visit happen, and (2) \emph{what} will happen in the $(i+1)$-th visit.
We use the decoder hidden state at timestep $i$, $\mathbf{s}_i$, as the latent vector of the $(i+1)$-th visit to make these predictions.
For disease prediction:
\begin{equation}\small
    \label{eq:disease_pred}
    \mathbf{\hat{x}}_{i+1} = Softmax(\mathbf{W}_o \mathbf{s}_i + \mathbf{b}_o)
\end{equation}
where $\mathbf{W}_o^T \in \mathbb{R}^{|\mathcal{C}| \times D}, \mathbf{b}_o \in \mathbb{R}^{|\mathcal{C}|}$,
and $D$ is the dimension of the decoder hidden state, $\mathbf{s}_i$.
We use categorical cross-entropy between the actual and the predicted codes as the loss for disease prediction.
\begin{equation}\small
    \label{eq:L_disease}
    \mathcal{L}_{d_{(i)}} = - \big ( \mathbf{x}_{\scriptsize{i+1}}^\top \cdot \log \ (\mathbf{\hat{x}}_{\scriptsize i+1}) + (1 - \mathbf{x}_{\scriptsize{i+1}})^\top \cdot \\log \ (\mathbf{1 - \hat{x}}_{\scriptsize{i+1}}) \big )
\end{equation}

To model the point process, a novel intensity function with temporal cascade relationships is defined as:
\begin{align}\small
    \lambda^{*}(t) &= \exp \Big( \underbrace{\mathbf{W}_h^\top \mathbf{h}_{i}}_{\text{explicit temporal influence}} + \underbrace{\mathbf{W}_s^\top \mathbf{s}_{i}}_{\text{implicit cascade influence}} +\nonumber\\
    & \underbrace{\mathbf{W}_u^\top \mathbf{v}_{i}}_{\text{current visit influence}} + \underbrace{W_t (t - t_i)}_{\text{current time influence}} + \underbrace{b_t}_{\text{base intensity}} \Big) 
    \label{eq:lambda}
\end{align}
where $\mathbf{W}_h,\mathbf{W}_s \in \mathbb{R}^{D \times 1}$, $\mathbf{W}_u \in \mathbb{R}^{D_m \times 1}$ and $W_t, b_t \in \mathbb{R}$. 
Using this intensity function, we define the likelihood that the next visit event will occur at a predicted time $t$ using the conditional density function~\cite{DBLP:conf/kdd/rmtpp}:
\begin{equation}\small
    p(t | \mathbf{x}_i, \mathbf{z}_i, \mathbf{s}_i) = f^{*}(t) = \lambda^{*}(t) \ \exp \big( - \int_{t_i}^{t} \lambda^{*}(\tau) d \tau \big)
\end{equation}

This likelihood of observing the next visit at the ground truth time $t_{i+1}$ in the training data should be maximised to learn the temporal cascade relationships.
Specifically, we minimise the negative log likelihood for time modelling as:
\begin{equation}\small
    \label{eq:L_time}
    \mathcal{L}_{t_{(i)}} = - \log \ \big ( p(t_{i+1} | \mathbf{x}_i, \mathbf{z}_i, \mathbf{s}_i) \big )
\end{equation}

Once the model is trained, to predict the time of the next visit at timestep $i+1$, we adopt the expectation formula~\cite{DBLP:conf/kdd/rmtpp}: 
\begin{equation}
    \hat{t}_{i+1} = \int_{t_i}^{\infty} t \cdot f^*(t) dt
    \label{eq:time_pred}
\end{equation}

\subsubsection{Overall Loss Function}
For $P$ patients we define the overall loss (Eqs.~\ref{eq:L_struct},~\ref{eq:L_disease},~\ref{eq:L_time}) with hyperparameters $\alpha,\beta>0$:
\begin{equation}\small
    \label{eq:L_all}
    \mathcal{L}(\{\mathcal{S}^{(p)}\}_{p=1}^{P}) = \frac{1}{P} \sum_{p=1}^{P} \bigg( \frac{\sum_{i=1}^{N^{(p)}} \big ( \mathcal{L}_{{d}_{(i)}}^{(p)} + \alpha \mathcal{L}_{{t}_{(i)}}^{(p)} \big ) + \beta \mathcal{L}_{s}^{(p)}}{N^{(p)}}  \bigg)
\end{equation}

\subsection{Model Optimisation}
\texttt{MEDCAS} jointly optimises both structural and temporal aspects in an end-to-end framework.
It connects the two parts, by feeding the visit embedding encoded by the structural part as the marker of the point process in the temporal part. 

For the structural part, a na\"{i}ve edge sampling method that is independent of the temporal sequence sampling will be completely random. 
More specifically, the structural part will select visit-code pairs that are distinct from the patients sampled by the temporal part in each minibatch.
Such distinct visit-code samples can deviate the model when updating the parameters in the optimisation phase.
To mitigate this issue, we propose a novel \textbf{visit-sequence-based edge sampling} method to ensure that the visit-code relations we learn are related to the sampled set of patients.
We first sample a batch of patients for the temporal part.
Then, for the structural part, we use the visits in these patients to sample the visit-code edges as the set of positive edges, thus removing the randomness of positive edge sampling to avoid cluttered joint training for the two components.
Given the sampled positive edges, we randomly select $K$ negative edges for each positive edge by fixing the visit node for the structural component.
This new sampling approach has been experimentally validated to be effective in our Ablation Analysis.

\section{Experiments}

We evaluate the performance of \texttt{MEDCAS} on three real-world EHR datasets and compare against several state-of-art EHR machine learning methods. 
Source code is available through this link: \url{https://bit.ly/3hvo0IV}.
All the experiments were performed on a MacBook Pro laptop with 16GB memory and a 2.6 GHz Intel Core i7 processor.

\subsection{Experimental Settings}

\subsubsection{Datasets}
\label{sec:datasets}

Two real-world proprietary cohort-specific EHR datasets: heart failure (HF) and chronic liver disease (CL), and one public ICU EHR dataset: MIMIC\footnote{\url{https://mimic.physionet.org}} are used. MIMIC is chosen to ensure the reproducibility on publicly available datasets. The two chronic diseases, HF and CL, are chosen as most patients with these disease tend to exhibit multimorbidity~\cite{almirall2013coexistence}.

We filter out the patients with only one visit.
Brief statistics of the three datasets are shown in Table~\ref{tab:data_stat}.
Both proprietary datasets are from the same hospital which use \textit{ICD-10 medical codes}.
MIMIC uses \textit{ICD-9 medical codes}.
Also, we can categorise the medical codes in accordance with the corresponding clinical classification software (CCS)\footnote{\url{https://hcup-us.ahrq.gov/toolssoftware/ccsr/ccs_refined.jsp}}, which divides the codes into clinically meaningful groups, to reduce the dimensionality in training (Eq.~\ref{eq:L_disease}).
We randomly split patient sequences in each dataset into train:validation:test with a 80:5:15 ratio.

\begin{table}[tb]
    \begin{center}
    \caption{Statistics of the real-world EHR datasets.
    }\label{tab:data_stat}
    \small
        \begin{tabular}{lrrr}
            \toprule
            \textbf{Dataset} & \textbf{HF} & \textbf{CL} & \textbf{MIMIC} \\
            \midrule
            Total \# patients & 10,713 & 3,830 & 7,499 \\
            Total \# visits & 204,753 & 122,733 & 19,911 \\
            Avg. \# visits per patient & 18.51 & 29.84 & 2.66 \\
            \midrule
            \multicolumn{4}{c}{\textbf{Medical Codes}}\\
            Total \# unique codes & 8,541 & 8,382 & 8,993 \\
            Avg.\ \# codes per visit & 5.55 & 5.05 & 16.92 \\
            Max \# codes per visit & 83 & 94 & 62 \\
            \midrule
            \multicolumn{4}{c}{\textbf{CCS Grouped Medical Codes}}\\
            Total \# unique codes & 249 & 248 & 294 \\
            Avg.\ \# codes per visit & 3.02 & 2.82 & 13.10 \\
            Max \# codes per visit & 33 & 37 & 14 \\
            \bottomrule
        \end{tabular}
    \end{center}
    \vspace{-4mm}
\end{table}

\subsubsection{Compared Algorithms}
Different state-of-art techniques on time and disease prediction are compared.

\textbf{Time prediction:}
Three state-of-the-art time models are compared.
\textbf{DoctorAI}~\cite{DBLP:conf/mlhc/doctorai} is a RNN-based encoder model which is able to predict both time and medical codes of the next visit by exploiting the chained memory structure in visits. \textbf{Homogeneous Poisson Process (HPP)}~\cite{DBLP:journals/siamrev/pp} is a stochastic point process model where the intensity function is a constant. It produces an estimate of the average inter-visit gaps. \textbf{Hawkes Process (HP)}~\cite{hawkes_process} is a stochastic point process model that fits the intensity function with an exponentially decaying kernel $\lambda^{*}(t) = \mu + \alpha \sum_{t_i < t} \exp (-\beta (t - t_i))$.
Considering time prediction as a regression task, we train a \textbf{linear regression model (LR)} and a \textbf{XGBoost regression model (XGBR)} with combination of input features: raw visit features $\mathbf{x}_i$ (i.e., medical code multi-hot vector at timestep $i$), visit embeddings $emb^{visit}_{i}$ (i.e., marker representation vector $\mathbf{v}_i$ at timestep $i$) and patient embeddings $emb^{pat}_{i}$ (i.e., decoder state vector $\mathbf{s}_i$ at timestep $i$) learned from \texttt{MEDCAS}.

\textbf{Disease prediction:}
Five state-of-the-art EHR methods are compared: \textbf{Med2Vec}, \textbf{DoctorAI}, \textbf{RETAIN}, \textbf{Dipole} \textbf{T-LSTM} and \textbf{MedGraph}, which have been discussed in the Related Work section. The main differences are summarized in Table~\ref{tab:baselines}.





\subsubsection{Evaluation Metrics}

For the time prediction task, root mean squared error (RMSE) is used as the evaluation metric. 
For the disease prediction task, Recall@k and AUC scores are used to evaluate model performance. Since the disease prediction task is a multi-label classification task, AUC score can be used to (micro-)average over all medical code classes. On the other hand, the disease prediction task can also be evaluated as an item ranking task as
we can rank the predicted codes in each visit based on the probability scores obtained with our model.
Recall@k computes the ratio of the number of relevant codes in the ranked top-$k$ predicted codes in the total number of actually present codes.

\subsubsection{Hyperparameter Settings}

We set the visit embedding dimension ($D_m$ in \texttt{MEDCAS}) in all the methods to $128$.
In our model, we fix $\alpha = 0.01$ and $\beta = 100$ in all three datasets based on the validation set.
We fix the number of negative codes randomly selected per positive code as $K=2$.
Based on the validation set, we fix the size of time-context vector embedding to be $8$ in HF, $64$ in CL and $128$ in MIMIC.
We use an Adam optimizer with a fixed learning rate of $0.001$.
For regularization in \texttt{MEDCAS} we use L2 norm with a coefficient of $0.001$.
We run each algorithm $100$ epochs with a batch size of $128$ and report their best performance across all the epochs.

\subsection{Time Prediction: \emph{When will it happen?}}

\begin{table}[tb]
  \caption{Time prediction performance. RMSE of log time. $\mathbf{x}_i$ is the medical code vector of the current visit, $emb^{visit}_{i}$ is the \texttt{MEDCAS} visit embedding vector and $emb^{pat}_{i}$ is the \texttt{MEDCAS} patient embedding vector (more details in Compared Algorithms).
  }
  \label{tab:time_prediction}
  \begin{center}
    \small
    \begin{tabular}{lccc}
    \toprule
    \textbf{Algorithm} & \textbf{HF} & \textbf{CL} & \textbf{MIMIC}\\
    \midrule
    HPP & 3.5916 & 3.5279 & 4.9734 \\
    HP & 3.2410 & 3.5613 & 4.4706 \\
    DoctorAI & 1.8312 & 1.9053 & 1.6273 \\
    \midrule
    LR ($\mathbf{x}_i$) & 8.2E+4 & 7.6E+9 & 4.8E+10 \\
    LR ($emb^{visit}_{i}$) & 2.3659 & 2.2590 & 1.5628 \\
    LR ($emb^{pat}_{i}$) & 2.0094 & 1.8663 & 1.4668 \\
    LR ($emb^{visit}_{i} + emb^{pat}_{i}$) & 2.0029 & 1.8579 & 1.4303 \\
    LR ($\mathbf{x}_i + emb^{visit}_{i}$) & 2.3702 & 2.2641 & 1.5431 \\
    LR ($\mathbf{x}_i + emb^{pat}_{i}$) & 2.0051 & 1.8593 & 1.3963 \\
    LR ($\mathbf{x}_i + emb^{visit}_{i} + emb^{pat}_{i}$) & 2.0060 & 1.8606 & 1.4079 \\
    \midrule
    XGBR ($\mathbf{x}_i$) & 2.3657 & 2.2609 & 1.5188 \\
    XGBR ($emb^{visit}_{i}$) & 2.3717 & 2.2679 & 1.5639 \\
    XGBR ($emb^{pat}_{i}$) & 2.0281 & 1.8909 & 1.4461 \\
    XGBR ($emb^{visit}_{i} + emb^{pat}_{i}$) & 2.0294 & 1.8844 & 1.4294 \\
    XGBR ($\mathbf{x}_i + emb^{visit}_{i}$) & 2.3709 & 2.2627 & 1.5328 \\
    XGBR ($\mathbf{x}_i + emb^{pat}_{i}$) & 2.0298 & 1.8962 & 1.4139 \\
    XGBR ($\mathbf{x}_i + emb^{visit}_{i} + emb^{pat}_{i}$) & 2.0265 & 1.8706 & 1.4172 \\
    \midrule
    \texttt{MEDCAS} & \textbf{1.7091} & \textbf{1.6950} & \textbf{1.3950} \\
    \bottomrule
  \end{tabular}
  \end{center}
  \vspace{-5mm}
\end{table}

We predict the log-scaled time interval between the current visit and the next visit and report the RMSE in Table~\ref{tab:time_prediction} for all three datasets. 
The lower the RMSE, the better the time prediction performance is.
Since the time gaps between consecutive visits of a patient can be highly irregular, the time to the next visit can be extremely skewed.
Therefore, similarly to DoctorAI~\cite{DBLP:conf/mlhc/doctorai}, we compute the RMSE in predicting the log scaled time value to mitigate the effect of the irregular time durations in the RMSE metric in all the compared models.

\texttt{MEDCAS} is the best in predicting \emph{when} the next visit occurs on all the datasets, by leveraging the temporal cascade relationship modelling and structural code-sharing behaviour modelling. 
The parametric point process models, HPP and HP, perform the worst because of the fixed parametric assumptions they make about the generative process (constant intensity of HPP and exponential kernel of HP). 
On the other hand, both sequential deep learning models (i.e., \texttt{MEDCAS} and DoctorAI) demonstrate superiority in predicting the time of next visit by learning the complex temporal relationships in the presence of additional information on visits, i.e., medical codes. 
\texttt{MEDCAS} has a very obvious improvements compared to DoctorAI, as DoctorAI only encodes the causal sequential ordering in a chained structure of visits using an RNN.
In contrast, \texttt{MEDCAS} models the varying influences of ancestor visits via point process modelling and captures the implicit temporal cascade relationships through attention.

Then, we compare the classic regression models (i.e., LR and XGBR) trained with different combinations of input features (i.e., the raw features, visit embeddings learned by \texttt{MEDCAS} and patient embeddings learned by \texttt{MEDCAS}).
Based on the results, there are two major observations, (1) the raw features ($\mathbf{x}_i$) and the visit embeddings ($emb^{visit}_{i}$) do not perform well in both LR and XGBR, and (2) the patient embedding ($emb^{pat}_{i}$) perform well in these two methods.
The patient embedding learned via \texttt{MEDCAS} has accumulated time information along with the longitudinal historical visits.
But, the raw features and the visit embeddings only have the medical code co-location information and do not capture the temporal information from the visit sequences.
Therefore, the results show that the inclusion of time context through the patient embedding improves the performance of time prediction task across all the datasets.
However, \texttt{MEDCAS} outperforms the two simple regression models.
This shows that the combined effect of structural and temporal learning with point process based time estimation using Eq.~\ref{eq:time_pred} in \texttt{MEDCAS} is effective.


\subsection{Disease Prediction: \emph{What will happen?}} 

\begin{table*}[tbh]
  \caption{Disease prediction performance.}
  \label{tab:disease_prediction}
  \begin{center}
    \small
    \begin{tabular}{lccccccccccccc}
    \toprule
    & \multicolumn{4}{c}{\textbf{HF}} & \multicolumn{4}{c}{\textbf{CL}} & \multicolumn{4}{c}{\textbf{MIMIC}}\\ 
    \cmidrule(lr){2-5}\cmidrule(lr){6-9}\cmidrule(lr){10-13}
    \textbf{Algorithm} & \multicolumn{3}{c}{\textbf{Recall@k}}  && \multicolumn{3}{c}{\textbf{Recall@k}} && \multicolumn{3}{c}{\textbf{Recall@k}} & \\ 
    \cmidrule(lr){2-4} 
    \cmidrule(lr){6-8}
    \cmidrule(lr){10-12}
    & \textbf{k=10} & \textbf{k=20} & \textbf{k=30} & \textbf{AUC} & \textbf{k=10} & \textbf{k=20} & \textbf{k=30} & \textbf{AUC} & \textbf{k=10} & \textbf{k=20} & \textbf{k=30} & \textbf{AUC}\\
    \midrule
    Med2Vec& 0.5735 & 0.7136 & 0.7945 & 0.9301 & 0.6414 & 0.7547 & 0.8232 & 0.9413 & 0.3017 & 0.4603 & 0.5666 & 0.8791 \\
    DoctorAI & 0.5006 & 0.6381 & 0.7132 & 0.7567 & 0.6011 & 0.7341 & 0.8117 & 0.8461 & 0.3026 & 0.4607 & 0.5659 & 0.7406 \\
    RETAIN & 0.5495 & 0.6867 & 0.7646 & 0.9214 & 0.6183 & 0.7319 & 0.7918 & 0.9281 & 0.2573 & 0.4135 & 0.5262 & 0.8643 \\
    Dipole & 0.5555 & 0.6942 & 0.7751 & 0.9254 & 0.6487 & 0.7610 & 0.8246 & 0.9403 & 0.3218 & 0.4819 & 0.5878 & 0.8781 \\
    T-LSTM & 0.3509 & 0.5554 & 0.6858 & 0.8275 & 0.3747 & 0.5477 & 0.7100 & 0.8508 & 0.2067 & 0.3878 & 0.5435 & 0.8279 \\
    MedGraph & 0.5514 & 0.6950 & 0.7793 & 0.9277 & 0.6601 & 0.7693 & 0.8331 & 0.9421 & 0.2998 & 0.4577 & 0.5671 & 0.8779 \\
    \midrule
    \texttt{MEDCAS} & \textbf{0.6118} & \textbf{0.7419} & \textbf{0.8107} & \textbf{0.9373} & \textbf{0.6832} & \textbf{0.7893} & \textbf{0.8445} & \textbf{0.9479} & \textbf{0.3350} & \textbf{0.4901} & \textbf{0.5928} & \textbf{0.8816} \\
    \bottomrule
  \end{tabular}
  \end{center}
  \vspace{-2mm}
\end{table*}

We report the Recall@k and AUC scores on the disease prediction task on the three datasets in Table~\ref{tab:disease_prediction}. 

As can be seen from the results, \texttt{MEDCAS} achieves the best performance on both recall and AUC across all three datasets, in particular showing significantly higher Recall@k for all $k$. \texttt{MEDCAS} achieves better performance over the existing attention-based encoder models (i.e., RETAIN and Dipole) due to two possible reasons. One reason is that RETAIN and Dipole learn attention by computing the relative importance of past visits on the current visit instead of the next visit. The other possible reason is that \texttt{MEDCAS} learn the attention more accurately compared with RETAIN and Dipole as it leverages both \textit{what} and \textit{when} aspects in modelling the patient sequences. 
Another observation is that the EHR graph structure learning methods, \texttt{MEDCAS} and MedGraph, perform well in disease prediction as they explicitly optimise for the code-sharing similarity of visits.

Moreover, disease prediction on MIMIC gives lower accuracy when compared with the other two. 
This is because of the substantially shorter visit sequences in MIMIC, which provides less information for a model to learn from about a patient (i.e., cold start patient problem).
Additionally, HF and CL are cohort-specific EHR datasets and the disease progression patterns of the patients will be fairly similar when compared to generic ICU admission in MIMIC.

\subsection{Performance Analysis on Cold Start Patients}

As shown in Figure \ref{fig:seq_lens_distribution}, many patients have short medical histories, with a median of 5 visits in HF and 1 visit in MIMIC. Intuitively, these short histories will make the accurate patient sequence modelling challenging. We refer this challenge as the ``\textbf{cold start patient problem}''.
In this section, we evaluate our model against the baselines to analyse their effectiveness in dealing with the cold start patient problem. 
We report the results on MIMIC and the results on HF and CL datasets have similar trends.

\begin{figure*}[htb]
    \centering
    \begin{minipage}{\linewidth}
        \centering
        \begin{subfigure}{0.29\linewidth}
            \centering
            \includegraphics[width=0.85\linewidth]{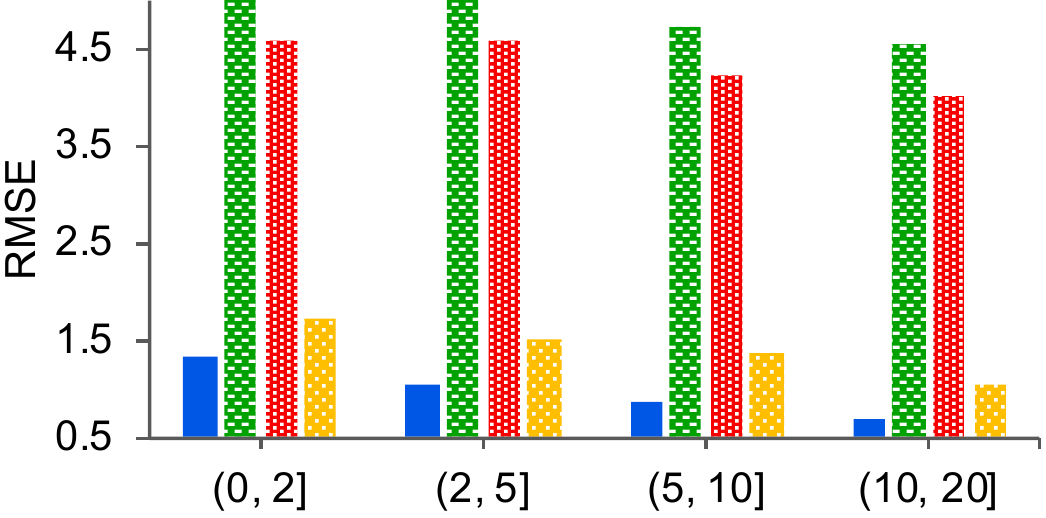}
            \caption{Time prediction (lower is better)}
            \label{fig:cs_mimic_time}
        \end{subfigure}
        \begin{subfigure}{0.65\linewidth}
            \centering
            \includegraphics[width=\linewidth]{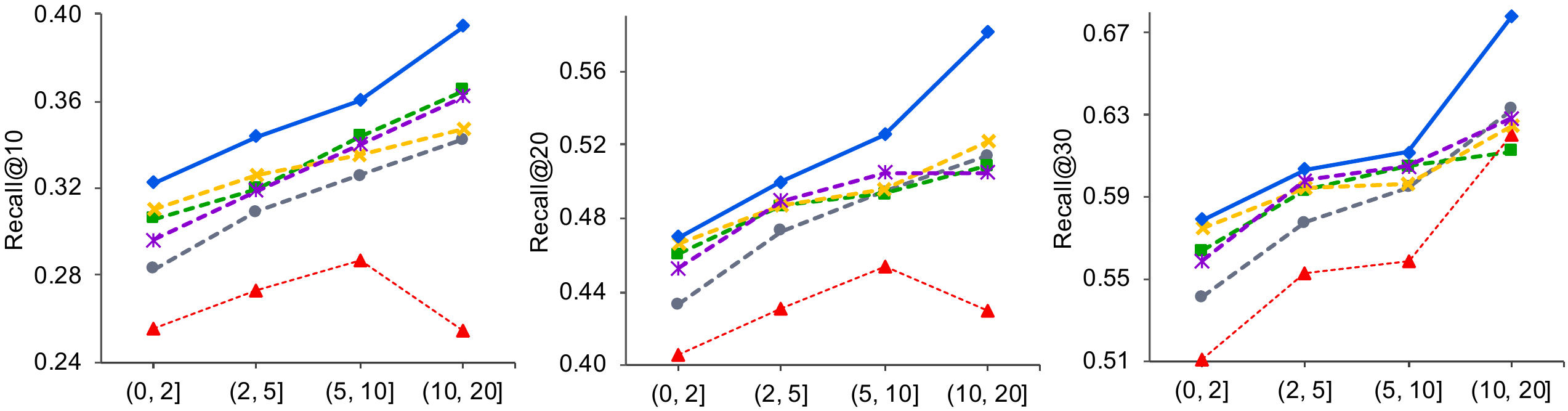}
        \caption{Disease prediction (higher is better)}
        \label{fig:cs_mimic_disease}
        \end{subfigure}
    \end{minipage}
    \caption{The cold start patient problem: time and disease prediction performance of cold start patient sequences measured (y-axis) in terms of patient sequence lengths (x-axis) in MIMIC.}
    \label{fig:cold_start}
    \vspace{-2mm}
\end{figure*}

For the patient sequences in the test set, we first remove the patients with only one visit and divide the patients into four partitions according to the length of patient sequences. 
Let $x$ be the length of patient sequences. 
We define the partitions as: $x \in (0, 2], (2, 5], (5, 10]$ and $(10, 20]$. 
We report the results on time and disease prediction in Figure~\ref{fig:cs_mimic_time}--~\ref{fig:cs_mimic_disease}. 

In both tasks, the longer the visit sequence is, the better the performance is for all the compared models. For the time prediction task, 
\texttt{MEDCAS} is performing the best for cold start patients as it can utilize the global code-sharing behaviour of visits of similar patients when the patient has short history. \texttt{MEDCAS} shows significantly better recall scores across all sequence partitions for disease prediction task.
Again, we can attribute this performance gain to the structure learning component, which  is further validated by the second-best performing model, MedGraph which also models the code-sharing structure.
RETAIN's performance is almost the worst as its reverse-time attention mechanism with two RNN structures.
When the sequences are shorter, RETAIN does not benefit from the reverse-time attention scheme and finds it difficult to infer a correct prediction.


\subsection{Ablation Analysis}
\label{sec:ablation}


\begin{table}[tb]
  \caption{Ablation analysis: time and disease prediction.
  }
  \label{tab:abl_disease_prediction}
  \begin{center}\small
    \begin{tabular}{lccccc}
    \toprule
    & \textbf{Time} & \multicolumn{4}{c}{\textbf{Disease}} \\
    \cmidrule{2-6} 
    \textbf{Variant} & & \multicolumn{3}{c}{\textbf{Recall@k}} & \\ 
    \cmidrule{3-5} 
    \textbf{version} & \textbf{RMSE} & \textbf{k=10} & \textbf{k=20} & \textbf{k=30} & \textbf{AUC} \\
    \midrule
    \texttt{MEDCAS}-1 & 1.404 & 0.3135 & 0.4709 & 0.5780 & 0.8775 \\
    \texttt{MEDCAS}-2 & 1.401 & 0.3105 & 0.4616 & 0.5641 & 0.8735 \\
    \texttt{MEDCAS}-3 & 1.405 & 0.3295 & 0.4831 & 0.5861 & 0.8790 \\
    \texttt{MEDCAS}-4 & 1.402 & 0.3276 & 0.4814 & 0.5812 & 0.8783 \\
    \texttt{MEDCAS}-5 & 1.400 & 0.3261 & 0.4832 & 0.5855 & 0.8792 \\
    \texttt{MEDCAS} & \textbf{1.395} & \textbf{0.3350} & \textbf{0.4901} & \textbf{0.5928} & \textbf{0.8816} \\
    \bottomrule
  \end{tabular}
  \end{center}
  \vspace{-6mm}
\end{table}

An extensive ablation study has been conducted to evaluate the effectiveness of each proposed component in \texttt{MEDCAS} on both disease and time prediction tasks (Table~\ref{tab:abl_disease_prediction}). 
Due to the limited space, the ablation study is only reported on MIMIC.
Five variant versions of \texttt{MEDCAS} are compared.

\textbf{\texttt{MEDCAS}-1}: without cascade modelling. 
It assumes that each visit is only affected by the immediate previous visit and no attention is learned.
\textbf{\texttt{MEDCAS}-2}: without visit-code graph structural modelling. It uses a jointly learned feedforward network to encode the multi-hot visit vectors ($\mathbf{x}_i$) into a latent marker.
\textbf{\texttt{MEDCAS}-3}: without time-context vector modelling. 
It directly concatenates the log-scaled time gap value $\Delta_i$ with the latent marker vector as the input to the sequence encoder.
\textbf{\texttt{MEDCAS}-4}: without visit-sequence-based edge sampling, where the model randomly samples positive edges and $K$ negative edges for each positive edge.
\textbf{\texttt{MEDCAS}-5}: 
with a single-parent visit instead of multiple ancestor visits to approximate a classic Hawkes process. Classic Hawkes process models assume that an event is affected by a single parent event in the history. The attention score function is modified to $\theta_{i,j} = \{0, 1\}, \sum_{j=1}^i \theta_{i,j} = 1$ so that only one past visit is chosen as the parent.

Based on the results reported in Table~\ref{tab:abl_disease_prediction}, we can see that \texttt{MEDCAS} full model outperforms all the variants on both tasks, which validate the effectiveness of the five components in our model. The consistent performance degradation for both tasks in \texttt{MEDCAS}-1 and \texttt{MEDCAS}-5 compared with \texttt{MEDCAS} validates the effectiveness of modelling temporal cascade relationships in patient visit sequences. Note that, although both \texttt{MEDCAS}-1 and \texttt{MEDCAS}-5 assume that each visit is affected by one single parent visit, \texttt{MEDCAS}-5 outperforms \texttt{MEDCAS}-1 consistently. The reason is that \texttt{MEDCAS}-5 learns to choose the parent visit from all the history visits while \texttt{MEDCAS}-1 limits the parent visit to be the immediate previous visit. Moreover,the consistent performance gap between \texttt{MEDCAS}-5 and \texttt{MEDCAS} validates that each visit can be affected by multiple previous visits. 

For time prediction, the performance gap between \texttt{MEDCAS} and \texttt{MEDCAS}-3 where time-context vector learning is removed, is the largest. This shows that time-context vector is very beneficial for time prediction as it adds an extra depth of information along the time axis. 
For disease prediction, the performance gap between \texttt{MEDCAS} and \texttt{MEDCAS}-2 where structure learning is removed is the largest. 
This demonstrates that the graph-based based code-sharing behaviour learning is very useful in this task. 
Moreover, the performance degradation of \texttt{MEDCAS}-1 is the second largest which shows the importance of temporal cascade modelling to account for multimorbidity in disease prediction.

\subsection{Parameter Sensitivity Analysis}

\begin{figure}
    \begin{minipage}{\linewidth}
        \centering
        \begin{subfigure}{0.49\linewidth}
            \centering
            \includegraphics[width=0.7\linewidth]{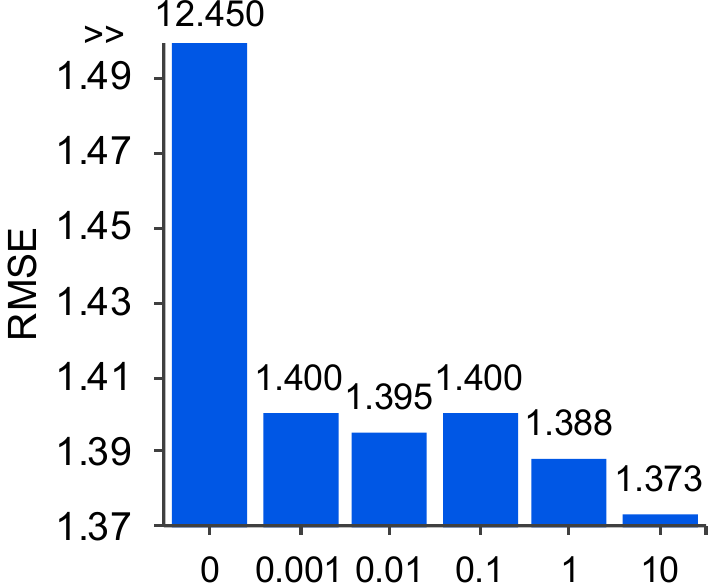}
            \caption{Time prediction with $\alpha$}
            \label{fig:ps_alpha_time}
        \end{subfigure}
        \begin{subfigure}{0.49\linewidth}
            \centering
            \includegraphics[width=0.7\linewidth]{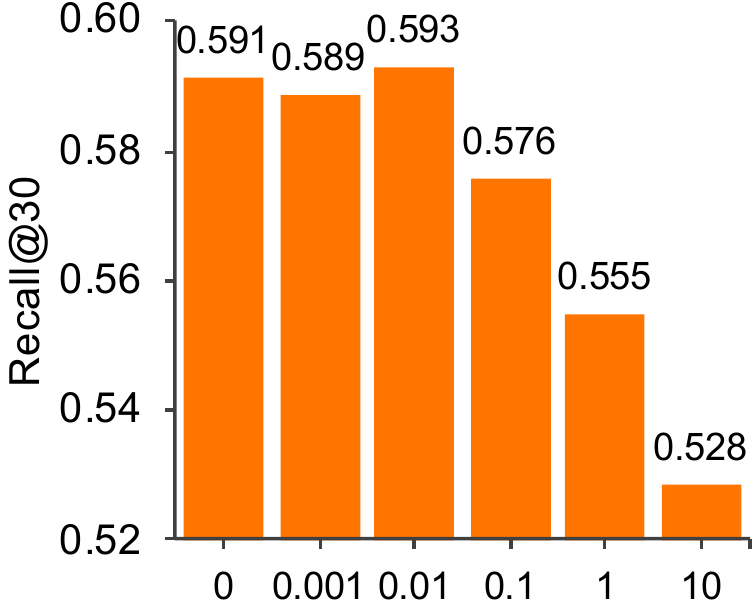}
            \caption{Disease prediction with $\alpha$}
            \label{fig:ps_alpha_disease}
        \end{subfigure}
    \end{minipage}
    \begin{minipage}{\linewidth}
        \centering
        \begin{subfigure}{0.49\linewidth}
            \centering
            \includegraphics[width=0.7\linewidth]{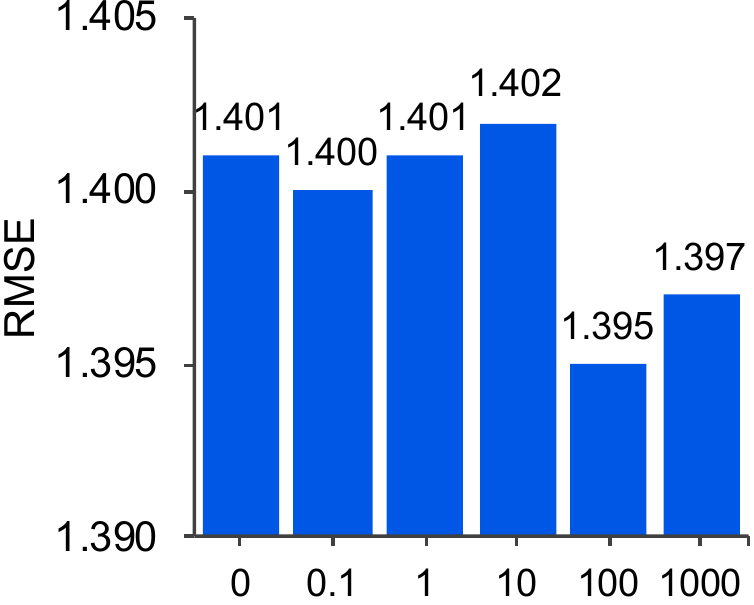}
            \caption{Time with $\beta$}
            \label{fig:ps_beta_time}
        \end{subfigure}
        \begin{subfigure}{0.49\linewidth}
            \centering
            \includegraphics[width=0.7\linewidth]{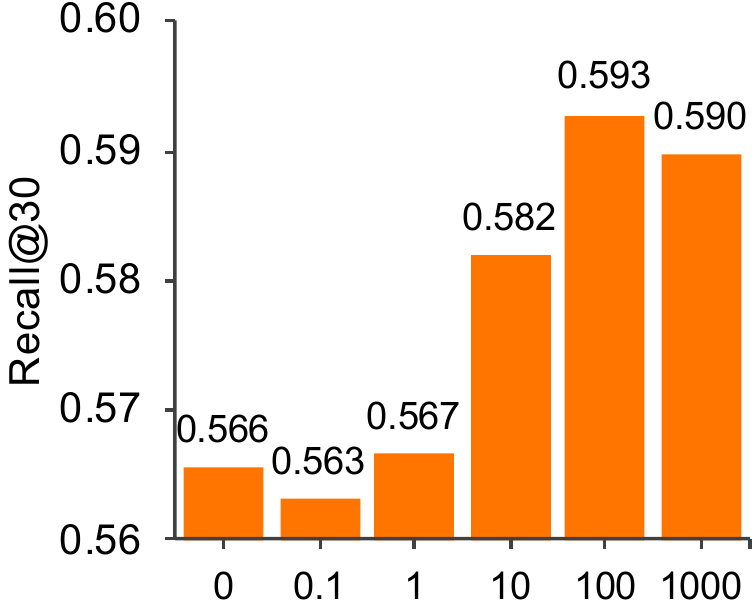}
            \caption{Disease with $\beta$}
            \label{fig:ps_beta_disease}
        \end{subfigure}
    \end{minipage}
    \begin{minipage}{\linewidth}
        \centering
        \hfill
        \begin{subfigure}{\linewidth}
            \centering
            \includegraphics[width=0.6\linewidth]{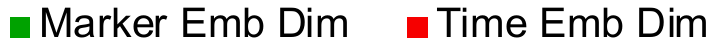}
        \end{subfigure}
        \begin{subfigure}{0.49\linewidth}
            \centering
            \includegraphics[width=1\linewidth]{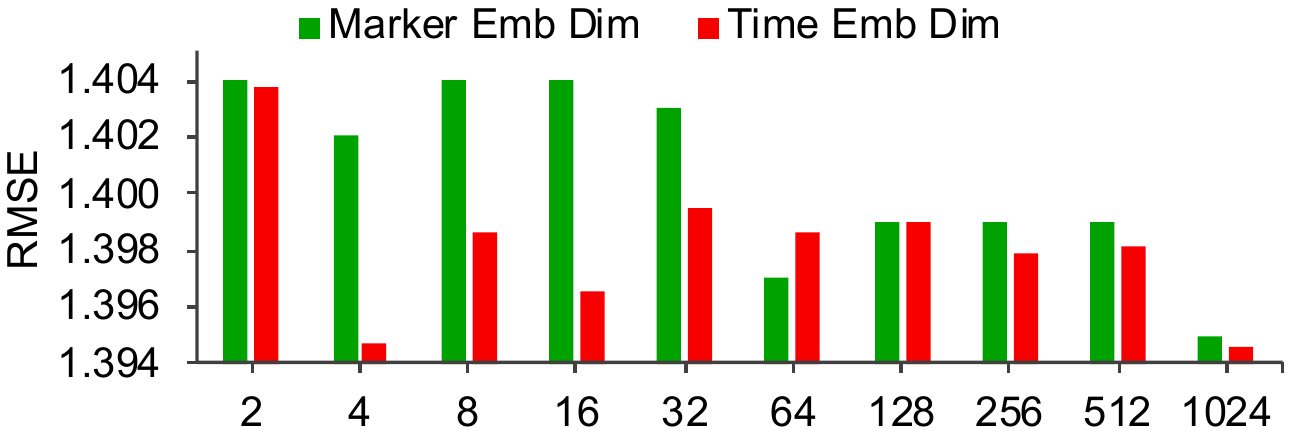}
            \caption{Time with emb dimension}
            \label{fig:ps_dim_time}
        \end{subfigure}
        \begin{subfigure}{0.49\linewidth}
            \centering
            \includegraphics[width=1\linewidth]{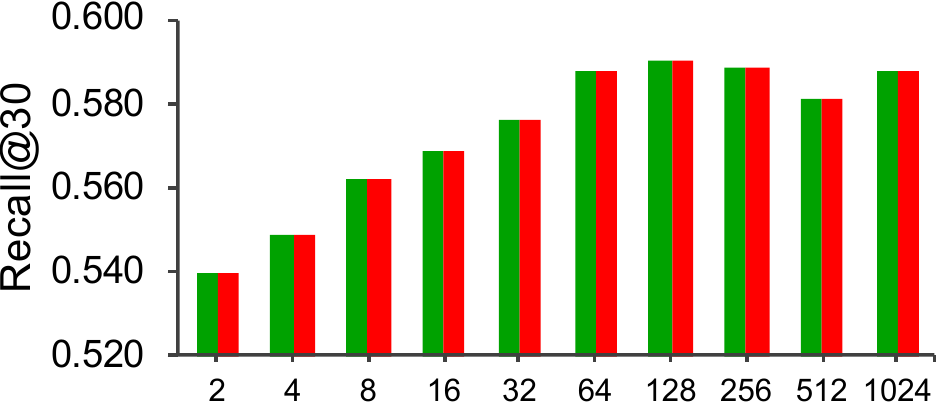}
            \caption{Disease with emb dimension}
            \label{fig:ps_dim_disease}
        \end{subfigure}
    \end{minipage}
    \caption{Parameter sensitivity. 
    Lower the better in Figs. (a), (c) and (e).
    Higher the better in Figs. (b), (d) and (f).}
    \label{fig:ps_analysis}
    \vspace{-5mm}
\end{figure}

We study the sensitivity of $\alpha$, $\beta$, marker embedding dimension and time-context embedding dimension in \texttt{MEDCAS}.
The results on time and disease prediction tasks on MIMIC are shown in Figure~\ref{fig:ps_analysis}.
Larger $\alpha$ values improve time prediction performance (Fig.~\ref{fig:ps_alpha_time}) but decrease disease prediction accuracy (Fig.~\ref{fig:ps_alpha_disease}), as $\alpha$ controls the relative weight of temporal learning as defined in Eq.~\ref{eq:L_all} (optimal at $\alpha = 0.01$).
When $\beta$ increases, both the time (Fig.\ref{fig:ps_beta_time}) and disease (Fig.\ref{fig:ps_beta_disease}) prediction performances improves, which demonstrates the contribution of the structural learning component in the two tasks (optimal at $\beta = 100$).

We analyse the trade-off of the marker embedding and the time embedding dimensions in Figure~\ref{fig:ps_dim_time} and Figure~\ref{fig:ps_dim_disease}.
In both settings, we keep the other dimension fixed at 128 and change the dimension under study.
The model performs increasingly well until 128 (resp.\ 128) in the disease prediction task with increasing marker (resp.\ time) embedding dimension while keeping the time (resp.\ marker) embedding fixed.
Time prediction performance also increases with marker (resp.\ time) embedding dimension and the best time accuracy is at 1024. 
Considering performance of both tasks, the optimal marker dimension at 128 and time-context dimension at 128 for MIMIC can be observed.

\section{Conclusion}
EHR data can be used for fine-grained readmission prediction in answering \emph{when} will the next visit occur and \emph{what} will happen in the next visit.
We presented \texttt{MEDCAS}, a novel temporal cascade and structural modelling framework for EHRs to simultaneously predict the time and the diseases of the next visit.
\texttt{MEDCAS} combines point processes and RNN-based models to capture multimorbidity condition in patients by learning the inherent time-gap-aware cascade relationships via attention in the visit sequences. 
The structural aspect of EHRs is leveraged with a graph structure based technique to alleviate the cold start patient problem in EHRs.
\texttt{MEDCAS} demonstrates significant performance gain over several state-of-the-art models in predicting time and diseases of the next visit across three real-world EHR datasets.

\bibliographystyle{IEEEtran}
\bibliography{bibliography}

\begin{thebibliography}{10}
\providecommand{\url}[1]{#1}
\csname url@samestyle\endcsname
\providecommand{\newblock}{\relax}
\providecommand{\bibinfo}[2]{#2}
\providecommand{\BIBentrySTDinterwordspacing}{\spaceskip=0pt\relax}
\providecommand{\BIBentryALTinterwordstretchfactor}{4}
\providecommand{\BIBentryALTinterwordspacing}{\spaceskip=\fontdimen2\font plus
\BIBentryALTinterwordstretchfactor\fontdimen3\font minus
  \fontdimen4\font\relax}
\providecommand{\BIBforeignlanguage}[2]{{%
\expandafter\ifx\csname l@#1\endcsname\relax
\typeout{** WARNING: IEEEtran.bst: No hyphenation pattern has been}%
\typeout{** loaded for the language `#1'. Using the pattern for}%
\typeout{** the default language instead.}%
\else
\language=\csname l@#1\endcsname
\fi
#2}}
\providecommand{\BIBdecl}{\relax}
\BIBdecl

\bibitem{DBLP:journals/titb/deep_ehr}
B.~Shickel, P.~Tighe, A.~Bihorac, and P.~Rashidi, ``Deep {EHR:} {A} survey of
  recent advances in deep learning techniques for electronic health record
  {(EHR)} analysis,'' \emph{{IEEE Journal of Biomedical and Health
  Informatics}}, vol.~22, no.~5, pp. 1589--1604, 2017.

\bibitem{goldstein2017opportunities}
B.~A. Goldstein, A.~M. Navar, M.~J. Pencina, and J.~Ioannidis, ``Opportunities
  and challenges in developing risk prediction models with electronic health
  records data: a systematic review,'' \emph{{JAMIA}}, vol.~24, no.~1, pp.
  198--208, 2017.

\bibitem{DBLP:journals/csur/ehr_survey}
P.~Yadav, M.~S. Steinbach, V.~Kumar, and G.~J. Simon, ``Mining electronic
  health records ({EHR}s): {A} survey,'' \emph{{ACM} Comput. Surv.}, vol.~50,
  no.~6, pp. 1--40, 2018.

\bibitem{journal/jama/readmission}
D.~Kansagara, H.~Englander, A.~Salanitro, D.~Kagen, C.~Theobald, M.~Freeman,
  and S.~Kripalani, ``{Risk Prediction Models for Hospital Readmission: A
  Systematic Review},'' \emph{{JAMA}}, vol. 306, no.~15, pp. 1688--1698, 2011.

\bibitem{DBLP:conf/mlhc/doctorai}
E.~Choi, M.~T. Bahadori, A.~Schuetz, W.~F. Stewart, and J.~Sun, ``Doctor {AI:}
  predicting clinical events via recurrent neural networks,'' in \emph{{MLHC}},
  2016.

\bibitem{DBLP:conf/ijcai/pacrnn}
Z.~Qiao, S.~Zhao, C.~Xiao, X.~Li, Y.~Qin, and F.~Wang, ``Pairwise-ranking based
  collaborative recurrent neural networks for clinical event prediction,'' in
  \emph{{IJCAI}}, 2018.

\bibitem{arxiv/medgraph}
B.~Hettige, W.~Wang, Y.~Li, S.~Le, and W.~L. Buntine, ``{MedGraph:} structural
  and temporal representation learning of electronic medical records,'' in
  \emph{{ECAI}}, 2020.

\bibitem{almirall2013coexistence}
J.~Almirall and M.~Fortin, ``The coexistence of terms to describe the presence
  of multiple concurrent diseases,'' \emph{Journal of {C}omorbidity}, vol.~3,
  no.~1, pp. 4--9, 2013.

\bibitem{journal/Navickas/Multimorbidity}
R.~Navickas, V.-K. Petric, A.~Feigl, and M.~Seychell, ``{Multimorbidity: What
  Do We Know? What Should We do?}'' \emph{{Journal of Comorbidity}}, vol.~6,
  no.~1, pp. 4--11, 2016.

\bibitem{journal/Afshar/Multimorbidity}
S.~Afshar, P.~Roderick, P.~Lowal, B.~Dimitrov, and A.~Hill, ``{Multimorbidity
  and the inequalities of global ageing: a cross-sectional study of 28
  countries using the World Health Surveys},'' \emph{{BMC} {P}ublic {H}ealth},
  vol.~15, no.~1, p. 776, 2015.

\bibitem{DBLP:conf/kdd/dipole}
F.~Ma, R.~Chitta, J.~Zhou, Q.~You, T.~Sun, and J.~Gao, ``Dipole: Diagnosis
  prediction in healthcare via attention-based bidirectional recurrent neural
  networks,'' in \emph{{ACM} {SIGKDD}}, 2017.

\bibitem{DBLP:conf/nips/retain}
E.~Choi, M.~T. Bahadori, J.~Sun, J.~Kulas, A.~Schuetz, and W.~F. Stewart,
  ``{RETAIN:} an interpretable predictive model for healthcare using reverse
  time attention mechanism,'' in \emph{{NIPS}}, 2016.

\bibitem{hawkes_process}
A.~G. Hawkes, ``Spectra of some self-exciting and mutually exciting point
  processes,'' \emph{{Biometrika}}, vol.~58, no.~1, pp. 83--90, 1971.

\bibitem{DBLP:conf/kdd/rmtpp}
N.~Du, H.~Dai, R.~Trivedi, U.~Upadhyay, M.~Gomez{-}Rodriguez, and L.~Song,
  ``Recurrent marked temporal point processes: Embedding event history to
  vector,'' in \emph{{ACM} {SIGKDD}}, 2016.

\bibitem{DBLP:conf/icdm/GaoWWYGWT019}
J.~Gao, X.~Wang, Y.~Wang, Z.~Yang, J.~Gao, J.~Wang, W.~Tang, and X.~Xie,
  ``{{CAMP:} Co-Attention Memory Networks for Diagnosis Prediction in
  Healthcare},'' in \emph{{ICDM}}, 2019.

\bibitem{gao2019camp}
J.~Gao, X.~Wang, Y.~Wang, Z.~Yang, .~J. Gao, J.~Wang, W.~Tang, and X.~Xie,
  ``{CAMP:} co-attention memory networks for diagnosis prediction in
  healthcare,'' in \emph{{ICDM}}, 2019.

\bibitem{kddd_2020}
M.~Zhang, C.~R. King, M.~Avidan, and Y.~Chen, ``Hierarchical attention
  propagation for healthcare representation learning,'' in \emph{{ACM SIGKDD}},
  2020.

\bibitem{ehr_icdm19}
C.~{Yin}, R.~{Zhao}, B.~{Qian}, X.~{Lv}, and P.~{Zhang}, ``Domain knowledge
  guided deep learning with electronic health records,'' in \emph{ICDM}, 2019.

\bibitem{kdd_2020_risk}
J.~Luo, M.~Ye, C.~Xiao, and F.~Ma, ``{HiTANet:} hierarchical time-aware
  attention networks for risk prediction on electronic health records,'' in
  \emph{{ACM SIGKDD}}, 2020.

\bibitem{DBLP:conf/kdd/med2vec}
E.~Choi, M.~T. Bahadori, E.~Searles, C.~Coffey, M.~Thompson, J.~Bost,
  J.~Tejedor{-}Sojo, and J.~Sun, ``Multi-layer representation learning for
  medical concepts,'' in \emph{{ACM} {SIGKDD}}, 2016.

\bibitem{conf/aaai/GCT}
E.~Choi, Z.~Xu, Y.~Li, M.~W. Dusenberry, G.~Flores, Y.~Xue, and A.~M. Dai,
  ``Learning the graphical structure of electronic health records with graph
  convolutional transformer,'' in \emph{{AAAI}}, 2020.

\bibitem{DBLP:conf/nips/word2vec}
T.~Mikolov, I.~Sutskever, K.~Chen, G.~S. Corrado, and J.~Dean, ``Distributed
  representations of words and phrases and their compositionality,'' in
  \emph{{NIPS}}, 2013.

\bibitem{DBLP:conf/kdd/tlstm}
I.~M. Baytas, C.~Xiao, X.~Zhang, F.~Wang, A.~K. Jain, and J.~Zhou, ``Patient
  subtyping via time-aware {LSTM} networks,'' in \emph{{ACM} {SIGKDD}}, 2017.

\bibitem{DBLP:conf/nips/mime}
E.~Choi, C.~Xiao, W.~F. Stewart, and J.~Sun, ``{MiME:} multilevel medical
  embedding of electronic health records for predictive healthcare,'' in
  \emph{{NeurIPS}}, 2018.

\bibitem{DBLP:conf/aaai/concare}
L.~Ma, C.~Zhang, Y.~Wang, W.~Ruan, J.~Wang, W.~Tang, X.~Ma, X.~Gao, and J.~Gao,
  ``{ConCare:} personalized clinical feature embedding via capturing the
  healthcare context,'' in \emph{{AAAI}}, 2020.

\bibitem{DBLP:conf/nips/neur_tpp}
H.~Mei and J.~Eisner, ``The neural hawkes process: {A} neurally self-modulating
  multivariate point process,'' in \emph{{NIPS}}, 2017.

\bibitem{DBLP:conf/ijcai/cascade_dyn}
Y.~Wang, H.~Shen, S.~Liu, J.~Gao, and X.~Cheng, ``Cascade dynamics modeling
  with attention-based recurrent neural network,'' in \emph{{IJCAI}}, 2017.

\bibitem{g2g}
A.~Bojchevski and S.~G{\"{u}}nnemann, ``Deep gaussian embedding of attributed
  graphs: Unsupervised inductive learning via ranking,'' in \emph{{ICLR}},
  2018.

\bibitem{DBLP:conf/www/line}
J.~Tang, M.~Qu, M.~Wang, M.~Zhang, J.~Yan, and Q.~Mei, ``{LINE:} large-scale
  information network embedding,'' in \emph{{WWW}}, 2015.

\bibitem{DBLP:journals/siamrev/pp}
P.~Guttorp, ``{An Introduction to the Theory of Point Processes {(D.} J. Daley
  and D. Vere-Jones)},'' \emph{{SIAM} Review}, vol.~32, 1990.

\end{thebibliography}

\end{document}